\documentclass[journal, onecolumn]{IEEEtran}

\usepackage[utf8]{inputenc}
\usepackage[T1]{fontenc}
\usepackage[english]{babel}

\usepackage{graphicx}
\usepackage{subfigure}
\usepackage[table]{xcolor}
\usepackage[justification=centering]{caption}

\usepackage{tabularx} 
\usepackage{array} 
\usepackage{booktabs} 
\usepackage{multirow}
\usepackage{multicol}
\usepackage[table]{xcolor}

\usepackage{mathtools}
\usepackage{amsmath} 
\usepackage{amssymb} 
\usepackage{amsthm} 

\usepackage{hyperref}
\usepackage{comment}
\usepackage{algpseudocode}
\usepackage{algorithm}
\usepackage{setspace}

\begin{document}
\title{Towards efficient compression and communication for prototype-based decentralized learning}
\author{Pablo Fernández-Piñeiro, Manuel Ferández-Veiga, Rebeca P. Díaz-Redondo, Ana Fernández-Vilas, Martín González-Soto
\thanks{Pablo Fernández-Piñeiro (iclab@det.uvigo.es), Manuel Fernández-Veiga (mveiga@det.uvigo.es), Rebeca P. Díaz-Redondo (rebeca@det.uvigo.es), Ana Fernández-Vilas (avilas@det.uvigo.es) and Martín González-Soto ( ) are with atlanTTic - I\&C Lab - Universidade de Vigo, Universidade de Vigo; Vigo, 36310, Spain.}
\thanks{Martín González-Soto (mgsoto@gradiant.org) is with Centro Tecnolóxico de Telecomunicacións de Galicia (GRADIANT),  Carretera do Vilar, 56-58, 36214, Vigo, Spain}}

\maketitle

\begin{abstract}
 In prototype-based federated learning, the exchange of model parameters between clients and the 
    master server is replaced by transmission of prototypes or quantized versions of the 
    data samples to the aggregation server. A fully decentralized deployment of prototype-
    based learning, without a central agregartor of prototypes, is more robust upon network 
    failures and reacts faster to changes in the statistical distribution of the data, 
    suggesting potential advantages and quick adaptation in dynamic learning tasks, e.g., 
    when the data sources are IoT devices or when data is non-iid.  In this paper, we 
    consider the problem of designing a communication-efficient decentralized learning 
    system based on prototypes. We address the challenge of prototype redundancy by 
    leveraging on a twofold data compression technique, i.e., sending only update messages 
    if the prototypes are information-theoretically useful (via the Jensen-Shannon 
    distance), and using clustering on the prototypes to compress the update messages used 
    in the gossip protocol. We also use parallel instead of sequential gossiping, and 
    present an analysis of its age-of-information (AoI). Our experimental results show that, with 
    these improvements, the communications load can be substantially reduced without 
    decreasing the convergence rate of the learning algorithm. 
\end{abstract}

\begin{IEEEkeywords}
Prototype-based learning,  Decentralized learning, Gossiping, IoT
\end{IEEEkeywords}


\section{Introduction}
\label{sec:Introduction}

Federated Learning (FL)~\cite{yang2019federated,kairouz2019advances,li2020federated} and Decentralized Federated Learning
(DFL)~\cite{beltran2023decentralized,li2020privacy} provide good approaches for
distributed machine learning system where the main focus is the minimization of a 
global loss function using different versions of a model created by multiple clients. 
These approaches have been extensively studied in the literature and applied, 
traditionally, to process private data in areas such as health and banking. In this 
paper, differently to these well-known approaches, we focus on the analysis and  
implementation of a decentralized machine learning system based on prototypes. 

On the one hand, our choice of prototype-based algorithms is motivated by the
advantages of these prototypes as compact representation of the data,
capturing the essential features and patterns within the dataset. This
property is particularly beneficial when reducing the complexity and volume of
data is crucial, such as in IoT systems and real-time applications, where it is usual 
to have large streams of data coming from the sensing network and where it is also 
common to have resource-limited devices.  By using prototypes, it is possible to 
handle large datasets and provide robust models that are less susceptible to 
overfitting~\cite{shao2014prototype, biehl2016prototype}. 

On another hand, our choice of a decentralized learning architecture entails there is not 
any designated node acting as central aggregator. Instead, all the nodes behave identically 
with the same role, and each one can exchange its local model built from its private 
datasets with some of their neighbors to update their models via a merging operation. Thus, 
the datasets (each one managed by one computation node or neighbor) are distributed across 
separate locations, and there is not any explicit transmission of the datasets to either a 
central aggregator or directly between the data owners. This scenario is specially 
appropriate for dynamic learning networks (devices on the move) and when the quality of the 
connections cannot always be guaranteed. Besides, by avoiding a central aggregator, we also 
avoid a single point of failure, which enhances the scalability and the dynamism of the 
learning setting.

Therefore, we focus on a decentralized learning architecture where the goal is to 
synthesize a good prototype-based supervised classification algorithm. This is a 
convenient approach to train models with non-iid data, i.e., with heterogeneous 
datasets at the clients or non-stationary data, since the calculation of the prototypes 
can be constrained to a specific timescale~\cite{Qiao2023}.

More specifically, we assume that the nodes in our system are fully connected in a
complete graph and each node runs an instance of Incremental Learning Vector Quantization 
(ILVQ~\cite{xu2012incremental, gonzalez2024decentralized}) for learning from its private 
dataset. ILVQ is a variant of LVQ~\cite{Kohonen1997} that adapts the set of prototypes 
(also known as dictionary or local model), using new input samples one by one to prune from 
the dictionary old prototypes that are no longer well suited to the statistical 
distribution of the recent inputs, and also for adding quickly new prototypes that  are 
similar to these fresh input values. Further, in our system, the nodes running ILVQ do not 
rely on a central master node for averaging or merging the local dictionaries. Instead, we 
assume that the nodes exchange asynchronously their dictionaries with a  random subset of 
neighbors, uniformly chosen according to a gossip protocol. This combination, a dynamic 
algorithm for classification and a gossip protocol, not only supports a decentralized 
learning structure but also inherently improves the system scalability and fault tolerance. 
Simultaneously, dissemination of information by means of gossiping helps to reduce latency 
in the learning process, as shown in~\cite{boyd2006randomized, kempe2003gossip}. Other 
advantages of decentralized learning via gossip techniques are the possible reduction
communication overhead~\cite{nedic2018distributed, giaretta2019gossip,hegedHus2019gossip}.
By contrast, a decentralized solution also faces challenges such as potential slower 
convergence times compared to centrally coordinated learning 
systems~\cite{boyd2006randomized}, or data inconsistencies across nodes due to  
asynchronous updates, requiring frequent exchanges to achieve model 
uniformity~\cite{shah2009gossip}. 

Precisely, our contributions address a problem that could reduce the efficiency of 
prototype-based learning. We aim to minimize prototype redundancy. This happens when 
training a local model with less prototypes attains better performance than training with a 
larger number of prototypes, and is due to two reasons: (i) the similarity among many 
prototypes, intra- or inter-class; (ii) the low quality of some prototypes, which can 
contain little new information about the features for updating a node's local model. 

 With this aim, our proposal consists of including several techniques to (i) to avoid unnecessary model exchanges and, (ii) to take advantage of the information received without reducing the prediction accuracy of the learning algorithm. This allows to enhance the scalability of the global system, specially for an IoT setting with resource-limited devices. Consequently, our contributions are the following ones:
\begin{itemize}
\item We define two mechanisms to reduce the number and size of the messages transmitted during gossiping to neighbor nodes. First, we propose to share only those model updates that contain relevant information that the neighbor nodes can use to improve their local models. This relevance is calculated using a variation of the Jensen-Shannon Distance~\cite{lin1991jsdistance}. Second, we propose to compress the local models before transmission. With this aim, we apply a density-based clustering algorithm~\cite{ester1996density} to reduce the prototype number without reducing the global model performance.
\item  We define an optimized scheduler to manage the incoming data to the computing nodes, that reduces the amount of information (local intermediate models from other neighbors) kept at each computing node and optimizes the training avoiding to train with potentially obsolete data.
\end{itemize}

In order to validate our proposal, we have followed a twofold procedure. On the one hand, we analyzed the dynamics of the gossip algorithm used in our proposal with the aim to obtain a lower bound for the freshness of the model computed locally at the nodes. On another hand, we have performed different experimental tests using resource-limited devices to measure the impact of the previously mentioned mechanisms to reduce the communication and computation overheads in this decentralized learning scenario and their impact in the performance of the prototype-based algorithm. 

The paper is organized as follows. In Sect.~\ref{sec:relatedWork} we overview 
interesting results in the literature related to decentralized learning and prototype-based algorithms. In Sect.~\ref{sec:background} we summarize relevant background knowledge related to our proposal, which is detailed in Sect.~\ref{sec:methodology}. To assess our approach we combine an mathematical analysis, which is detailed in 
Sect.~\ref{sec:analysis} with an experimental setting in Sect.~\ref{sec:validation}, where the obtained results are discussed and analyzed. Finally, conclusions and future work is detailed in Sect.~\ref{sec:conclusions}.

\section{Related Work}
\label{sec:relatedWork}

Decentralized learning based on gradient exchange has been previously studied in 
the literature as a solution to avoid the dependence on a critical central 
server, e.g. in~\cite{Yuan2023,MartinezBeltran2023}, both for robustness against 
failures and for better protection against malicious attacks. Additionally to 
these significant advantages, these works also highlight that learning in 
distributed/decentralized environments entails a higher communication overhead 
for the dissemination of the updates, so efficient forms of information 
dissemination (gossip) have been analyzed to reduce this cost. For instance, 
\cite{esposito2017improving} presents an approach that enhances gossiping protocols 
through distributed strategic learning. The authors apply tools from dynamic game 
theory to model the decision-making process among nodes, where each node independently 
selects which neighbors to communicate with to maximize network efficiency. A 
gossip-based distributed stochastic bandit algorithm is presented 
in~\cite{szorenyi2013gossip}, which formulates a stochastic multi-armed bandit 
strategy on peer-to-peer networks where each peer independently communicates with 
randomly selected peers.  Building on the previous strategies for optimizing network 
efficiency, \cite{hashemi2022benefits} investigates the impact of taking multiple 
gossip steps between nodes before performing the next local update in DFL. This 
study shows in particular that multiple compressed gossip steps (as opposed to a 
single high-precision gossip step) can significantly enhance the convergence rate 
of the learning process while staying within a fixed communication cost. Thus, more
frequent but lower-precision gossiping helps bridge the gap between centralized and 
decentralized convergence rates in gradient-based algorithms.

The convergence properties in a fully decentralized learning can also be altered 
because of the age of information (AoI) in the received model updates, i.e., the 
freshness of the incoming updates from neighbor nodes. Designing gossip protocols
with low expected AoI is key for attaining fast global learning when data are non-iid,
especially. The fundamental insights and models for understanding AoI and the trade-off
between its average value and the communication cost can be found 
in~\cite{yates2021age_gossip,mitra2024scale}, for instance. These works present 
analytical results useful to calculate the staleness of the message update process,
with a focus on the scaling laws for large systems. This is of interest for
characterizing the optimal frequency of updates so that the gossip mechanism can keep
the models updated while not saturating the neighbors with excessive (and redundant)
information. In particular, \cite{mitra2024scale} establishes that a minimum of 
$O(\log n)$ of messages have to be exchanged, on average for large $n$, in a network
with $n$ learning nodes in order to maintain a bounded AoI, and that at least $O(n)$
messages are necessary in opportunistic gossiping protocols. Most importantly, this
work also shows that failing to sustain these rates, the convergence of the 
decentralized learning algorithm breaks down.

For prototype-based learning, mitigating the problem of prototype redundancy has been
discussed in several works too through instance selection and prototype generation
techniques, as in~\cite{bezdek2001nearest}, that provides a comprehensive evaluation 
of multiple prototype generation  methods, including vector quantization, 
clustering-based approaches, and combinatorial search techniques. More dynamic solutions
for prototype management have been proposed, such as the sliding window 
technique~\cite{yang2012mining}, which purges outdated data in k-NN classifiers,
or the merge-purge approach introduced by~\cite{mccallum2000efficient}, that applies
clustering to high-dimensional data, merging similar instances to avoid redundancy. 
These methods adaptively manage prototypes to maintain efficiency in scenarios with large, 
continuous data streams

While most of the aforementioned studies consider either prototype-based FL or
decentralized versions of gradient-based learning, our work here goes into a 
relatively unexplored area: prototype-based decentralized learning with adaptive 
vector quantization at the nodes, specifically ILVQ. We handle the prototype
redundancy problem by means of a combination of techniques, jointly in operation.
First, a clustering-based method inspired by~\cite{garcia2016bigdata} is used to cut
the size of the model updates and control the volume of transmitted data. It 
also helps to reduce the computational load on the nodes during the continuous 
training. Secondly, the problem of low quality in the updates is handled by replacing
the usual metric with an estimate of the Jensen-Shannon distance. Our method for 
computing the distance, however, differs from the method discussed 
in~\cite{hoyos2023representation} in that we seek an efficient kernel-based algorithm
for inferring the pdf of the prototypes quickly. In all these respects, the present paper
extends and improves on our previous work~\cite{gonzalez2024decentralized}, where
decentralized learning with ILVQ was analyzed only with the goal of optimizing the
dynamic properties of ILVQ to process data streams.

\section{Background}
\label{sec:background}

\begin{table}[t]
  \centering
  \caption{Notation used in the ILVQ algorithm.} 
  \label{table:notation}
  \begin{tabular}{cp{0.8\textwidth}}
    \hline
    \textbf{Notation} & \textbf{Description} \\
    \hline
    $N$ & Number of total nodes in the current scenario \\
    $n_i$ & Unique identifier for each node \\
    $M_i$ & Local model in node $i$, $i=0,..., N-1$ \\
    $G_i$ & Prototype set of $M_i$ \\
    $s$ & Size of the subset of neighbors to share information with, constant for all nodes \\
    $x$ & New input pattern \\
    $y$ & Label of the new input pattern \\
    $E$ & Set storing connections between prototypes \\
    $T$ & Threshold for significant distance from a prototype \\
    $s_1$ & Winner prototype closest to $x$ \\
    $s_2$ & Runner-up prototype closest to $x$ \\
    \hline
  \end{tabular}
\end{table}

\begin{algorithm}[p]
    \setstretch{1.0}
  \caption{Simplified Incremental Learning Vector Quantization (ILVQ)}
  \label{alg:ILVQ}
  \begin{algorithmic}[1]
    \State Initialize prototype set $G$ with the first two input data from the training set
    \State Initialize a set $E$ to store connections between prototypes (empty initially)
    \Loop
    \State Receive new input pattern $x$ (with label $y$)
    \State Find the winner ($s_1$) and runner-up ($s_2$) in $G$ based on the smallest distance to $x$
    \If{$x$ represents a new class $\lor$ is significantly distant from $s_1$ $\lor$ $s_2$ based on threshold $T$}
    \State Add $x$ to $G$
    \State Proceed to process a new input pattern
    \EndIf
    \If{there's no connection between $s_1$ and $s_2$ in $E$}
    \State Add the connection ($s_1$, $s_2$) to $E$
    \State Set the age of this connection to 0
    \EndIf
    \State Increment the age for all connections involving $s_1$
    \State Increment the count of points for the class of $x$
    \If{the label of $s_1$ matches $x$}
    \State Move $s_1$ closer to $x$
    \For{each neighbor of $s_1$ with a different label}
    \State Move it away from $x$
  \EndFor
    \Else
    \State Move $s_1$ away from $x$
    \For{each neighbor of $s_1$ with the same label}
    \State Move it closer to $x$
    \EndFor
    \EndIf
    \State Remove old connections from $E$ based on age criteria
    \State Periodically remove nodes in $G$ without neighbors or in low-density areas
    \EndLoop \Comment{until all input patterns are processed}
  \end{algorithmic}
\end{algorithm}

As it was previously mentioned, we assume a totally decentralized learning setting where all the computing nodes are fully connected  in a complete graph and run an instance of Incremental Vector Learning Quantization (ILVQ). This algorithm is an adaptive, incremental version of the Learning Vector Quantization (LVQ)
algorithm~\cite{Kohonen1997}, a supervised learning classifier that uses a
sequence of labeled input vectors $(\mathbf{x}_i, c_i)$, with
$\mathbf{x}_i \in \mathbb{R}^m$, $c_i = 1, 2, \dots$ and class $i \in [C]$, to
calculate a small set of prototypes. A prototype $(\mathbf{p}_j, c_j)$ is
defined as a quantized version $\mathbf{p}_j = \phi(\mathbf{x}_{i_1}, \mathbf{x}_{i_2}, 
\dots, \mathbf{x}_{i_\ell}) \in \mathbb{R}^p$ of a subset of samples in a given 
class $c_j$. Here, $\phi(\cdot)$ denotes the feature extraction map from a subset 
of samples. Hence, the set of prototypes of class $c$ is
\begin{equation}
  P_c := \{ (\mathbf{p}_j, c_j) \in D : c_j = c \}
\end{equation}
where $D = \cup_{c \in [C]} P_c$ is the complete dictionary of
prototypes, also known as local model or intermediate model in decentralized learning jargon. Observe that a class $c$ can have multiple prototypes representing
the distribution of samples that belong to that class, and that the dictionary
is not static: new prototypes can be added as long as the algorithm learns
from the input stream of samples. 

ILVQ~\cite{xu2012incremental, gonzalez2024decentralized} is a variant of LVQ that adapts the dictionary $D$ incrementally, using new input samples one by one to update the dictionary of prototypes. ILVQ includes some rules to prune from the dictionary old prototypes that are no longer well suited to the statistical distribution of the recent inputs, and also for adding quickly new prototypes that are similar to these fresh input values. Therefore, ILVQ is robust, i.e., it is capable of tracking and self-learning the changes in the pdf of the input stream through continuous adaptation of the dictionary of prototypes. An outline of the ILVQ algorithm is included as Algorithm~\ref{alg:ILVQ}, for the sake of completeness. Refer to Table~\ref{table:notation} for the notation.
In our proposal, the nodes within the learning network exchange synchronously
their dictionaries $D_k$ with a random subset of $s$ neighbors, uniformly
chosen with probability $T$. Thus, this is a variant of the generic gossip protocol~\cite{hegedHus2019gossip}, detailed in Algorithm~\ref{alg:gossip_learning_protocol}.

\begin{algorithm}[t]
    \setstretch{1.0}
  \caption{Skeleton of the Gossip Learning Protocol}
  \label{alg:gossip_learning_protocol}
  \begin{algorithmic}[1]
    \Procedure{Main}{}
    \State $currentModel \gets \text{InitModel()}$
    \State $lastModel \gets currentModel$
    \Loop
    \State $\text{Wait}(\Delta)$
    \State $T \gets \text{RandomPeer()}$
    \State $\text{Send}(T, currentModel)$
    \EndLoop
    \EndProcedure
    
    \Procedure{OnModelReceived}{$m$}
    \State $currentModel \gets \text{UpdateMerge}(m, lastModel)$
    \State $lastModel \gets m$
    \EndProcedure
  \end{algorithmic}
\end{algorithm}

Nevertheless, since the message exchange protocol in our system has complexity
$O(s T n)$, where $n$ is the number of nodes, and the size of each individual
message can be potentially large, we propose to introduce some techniques oriented to 
reduce the communication and computation overheads, enhancing the scalability of the 
global system. One of them is based on sharing the intermediate model (local dictionary) 
only with the neighbors that may take advantage of this local knowledge. To this end, 
we propose to assess the novelty between both dictionaries by using the Jensen-Shannon
Distance~\cite{lin1991jsdistance}, that provides a symmetric and bounded measure of 
the difference between two distributions $P$ and $Q$
\begin{equation}
  \label{eq:jensen-shannon-distance}
  \operatorname{JS}(P, Q) = \sqrt{\frac{1}{2} D_{KL}(P || M) + \frac{1}{2}
    D_{KL}(Q || M)}, 
\end{equation}
where $M = \frac{P + Q}{2}$ and $D_{KL}(P || Q)$ is the divergence or
Kullback-Leibler distance
\begin{equation}
  \label{eq:divergence}
  D_{KL}(P || Q) = \sum_{i} P(x_i) \log \frac{P(x_i)}{Q(x_i)}.
\end{equation}
In our numerical experiments, as in other works as well~\cite{Yan2024}, we found that 
prototypes in the same class or in different classes can be quite similar, making it hard 
to distinguish between a new local prototype and a useful prototype for the global model. 
Furthermore, the $\ell_2$ distance used in other works~\cite{Qiao2023a} is unrelated to the 
information content of the features in the prototype, so it is not helpful to improve the 
learning speed. For this reason, we propose to supplement this distance with the JS 
distance, as a more appropriate metric to detect similarity between the empirical 
probability mass functions of two dictionaries of prototypes.

\section{Methodology}
\label{sec:methodology}

\begin{figure}[tpb]
    \centering
    \small
    \begin{tabular}{cl}
        \toprule
        \textbf{Notation} & \textbf{Description} \\
        \midrule
        $Q_{ij}$ & Queue of $M_i$ with the received $G_j$, $j \in \text{neighbors}_i$ \\
        $D_i$ & Dataset associated with node $M_i$ \\
        $D_{ik}$ & $k$-th sample of the dataset $D_i$\\
        $A_i = \{a_{ik}\}_{k=1}^{len(D_i)}$ & Set of arrival times at node $i$ \\
        $\lambda_s$ & Mean arrival rate of samples \\
        \bottomrule
        \vspace{0.1cm}
    \end{tabular}
    \includegraphics[width=0.75\textwidth]{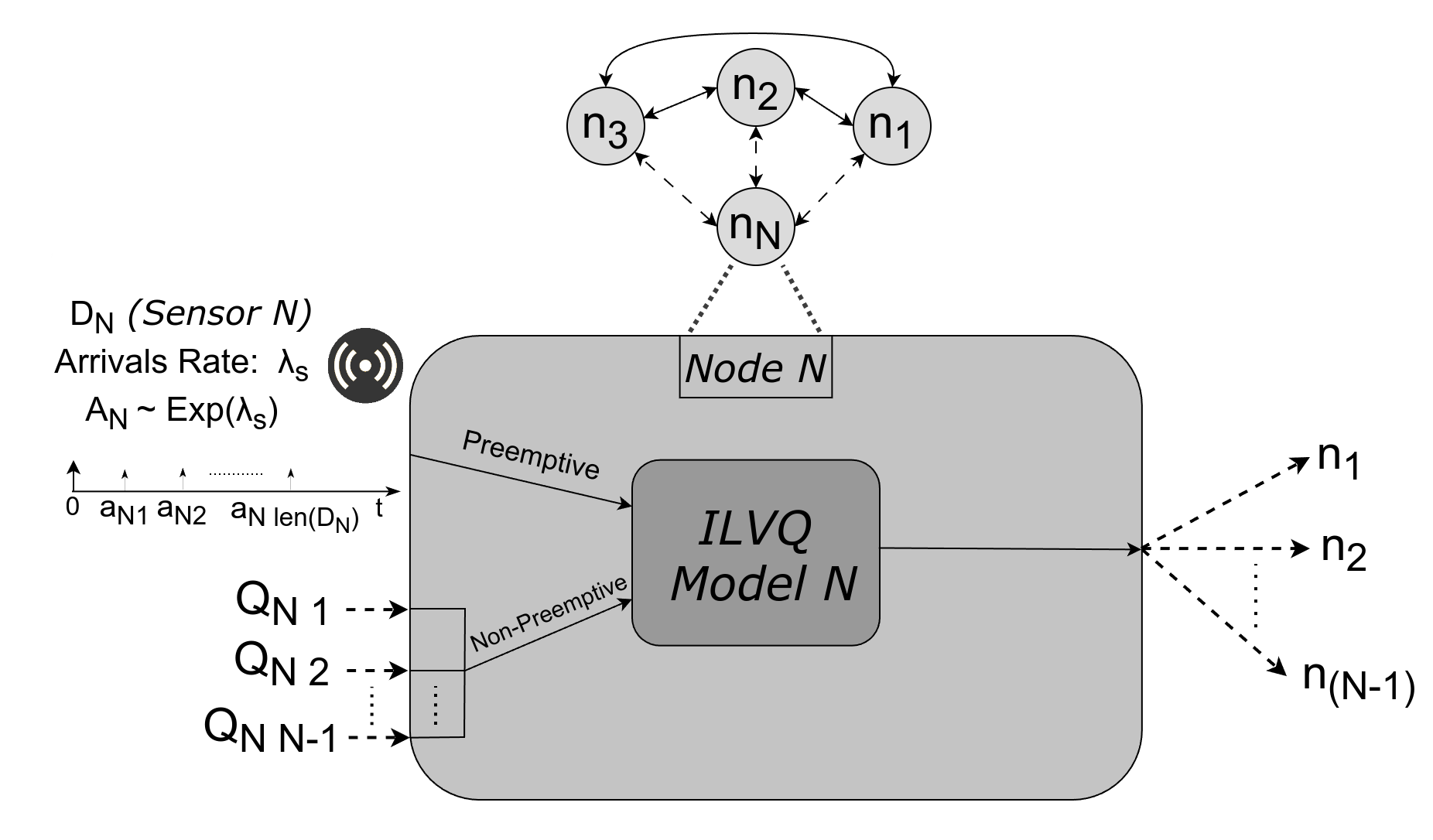}
    \vspace{0.1cm}
    \caption{Extended notations used in the decentralized learning framework and the general architecture of the simulated nodes.}
    \label{fig:architecture_and_notations}
\end{figure}

Our proposal, as it was previously mentioned, consists of a set of mechanisms to improve the efficiency of a totally decentralized learning scheme with limited-resources nodes. In this scenario, we consider a prototype-based learning algorithm, ILVQ~\cite{gonzalez2022xuilvq}, that individually runs at each one of the nodes in the learning network. These nodes share the intermediate model (dictionary or set of prototypes) with their peers using a gossip strategy in order to enrich the local knowledge with information from other nodes.

Fig.~\ref{fig:architecture_and_notations} depicts this scenario composed by $N$ nodes ($\{n_1, n_2, \ldots, n_N \}$), each one with its own sensing mechanism to gather data. These local samples ($D_{ik}$, sample $k$-th in dataset $D_i$) are modeled to arrive at the node sensing input with rate $\lambda_s$ and arrival times denoted as $a_{ik}$. With this information, each node $n_i$ creates its local model $M_i$, composed of a prototype set $G_i$. Additionally there is a queuing system composed of $N$ queues: one per node in the network ($N-1$), to handle the local models (dictionaries) sent by their peers; and another queue $0$ to receive the local data coming from the sensing devices. 

Each node in the decentralized learning network operates according to a three-stage cycle: (i) it trains its local incremental learning model (ILVQ) with the samples that receives from its sensor system ($D_{ik}$); (ii) it receives and processes local models from its peers ($M_j, j=1, \ldots, N, j\neq i$) to enrich its local learning model and (iii) shares, if appropriate, its own local model ($M_i$) within the network. 

In this Section, we describe in detail the three contributions of our proposal aligned with the objective of optimizing incremental learning solution for resource-limited devices. First, a scheduling mechanism to manage the queuing system where the local data (from sensors) and the intermediate models received from peers are temporally kept (Sect.~\ref{sec:queuing}). Then, a procedure to share the local models that is detailed in Sect.~\ref {sec:sharing}. Finally, a procedure to compress the prototype set in the local model that is detailed in Sect.~\ref {sec:clustering} 

\subsection{Learning process: optimized scheduler for the queuing system}
\label{sec:queuing}

\begin{algorithm}[tpb]
    \setstretch{1.0}
    \caption{Node -- learning model aggregation \& sharing}
    \label{alg:general}
    \begin{algorithmic}[1]
        \Procedure{Learning model aggregation \& sharing}{$A_i$, $D_{i}$}
            \State $currentQueueIndex \gets 0$
            \For{$a_{ik} \in A_i$}
                \While{waiting for $a_{ik}$}
                    \While{time not exceeded}
                        \State $currentQueueIndex \gets (currentQueueIndex + 1) \bmod (N - 1)$)  \Comment{Cyclic rotation}
                        \State LearnFromPeers($Q[currenQueueIndex]$)
                    \EndWhile
                \EndWhile
                \State LearnFromSensors($D_{i}$)            
                \State OptimizedSharingProtocol
            \EndFor
    \EndProcedure
    \end{algorithmic}
\end{algorithm}

\begin{algorithm}[tpb]
    \setstretch{1.0}
    \caption{LearnFromPeers}
    \label{alg:learn_from_queue}
    \begin{algorithmic}[1]
        \Procedure{LearnFromPeers}{$currentQueueIndex$}
             \If{$Q[currentQueueIndex]$ is not empty}
                \State $randomProtoIndex \gets \text{randint}(0, \text{len}(Q[currentQueueIndex]) - 1)$ 
                \State $prototype \gets Q[currentQueueIndex].pop(randomProtoIndex)$ \Comment{Pop a random item}
                \State ILVQ($prototype$) \Comment{Feed the ILVQ model}
                \State CompressModelSize($Model$) \Comment{Clustering is carried out in this step}
            \EndIf
        \EndProcedure
    \end{algorithmic}
\end{algorithm}

As the Algorithm~\ref{alg:general} shows, the fist action, processing the local sensor samples ({\texttt{LearnFromSensors}}, line $10$), is the priority for the node. Each time a new sensor sample arrives, the node reads and process this new value to train the incremental learning model. In the idle periods between sensor samples, the node is occupied doing the other two tasks: learning from other its peers' models and sharing its own local model. Consequently, there is a preemptive mechanism between the queue with sensing data and the queues where the peers' models are kept. 

Learning from the peers' models is done by processing the received ILVQ models, which are stored in the queuing scheme shown in Fig.~\ref{fig:architecture_and_notations}. These queues are served with a round-robin scheduling strategy with a quantum of one prototype per cycle (line 6 in  Algorithm~\ref{alg:general}). Once a queue is selected, the {\texttt{LearnFromPeers}} procedure (line 7 in  Algorithm~\ref{alg:general}) runs.

Since this methodology is defined for resource-limited devices, it is essential to manage (learn from) the incoming intermediate models in an efficient way to ensure the most recent data is never lost. For this, we have decided (i) to use a Last-In-First-Out (LIFO) strategy for each one of the queues; and (ii) to only store one set of prototypes per neighbor. This would guarantee the most recently received prototypes are processed, which is aligned with the adaptation to concept drift in dynamic environments. Additionally, to address situations where queues accumulate more prototypes than can be promptly processed, we apply a uniform random sampling method. That is, instead of processing the first few prototypes in a set, the system randomly selects an index for training, ensuring an uniform distribution across all received prototypes and enhancing the overall robustness and adaptability of the model (line $4$ in Algorithm~\ref{alg:learn_from_queue}). Consequently, this strategy supports an agile adaptation to changes while preventing resource waste on outdated data.

\subsection{Optimized Sharing protocol to minimize overheads}
\label{sec:sharing}

\begin{algorithm}[tpb]
    \setstretch{1.0}
        \caption{Optimized Sharing Protocol}
        \label{alg:RandomSharingProtocol}
        \begin{algorithmic}[1]
            \Procedure{OptimizedRandomSharing}{$s$, $T$, $N$}
                \State $shouldShare \gets \text{GenerateRandomNumber()} < T$  \Comment{Decide if sharing should occur}
                \If{$shouldShare$}
                    \State $availableneighbors \gets N - 1$ \Comment{Exclude self from neighbors}
                    \State $neighbors \gets \text{SelectRandomSubset}(availableneighbors, s)$ \Comment{Randomly select s neighbors}
                    \State $neighborsToShare \gets$ \text{IsItWorthy($neighbors$)?}          
                    \State \Return $neighborsToShare$
                \Else
                    \State \Return \{\} \Comment{No sharing if the decision is negative}
                \EndIf
            \EndProcedure
        \end{algorithmic}
    \end{algorithm}
    
One of the objectives of this methodology is to reduce the exchange of local models within the learning network to the minimum to support a correct operation of the ILVQ algorithm. Consequently, the local models are shared only when the information is interesting for the peers, i.e., increases their knowledge.

This procedure ({\texttt{OptimizedSharingProtocol}}: line 11 in Algorithm~\ref{alg:general}), is run by each one of the computing nodes and it works in two phases, as detailed in Algorithm~\ref{alg:RandomSharingProtocol}. First, each node randomly selects a set of peers to which potentially share the local model (line $5$). This is done applying a random sharing protocol based on the gossip algorithm philosophy~\cite{hegedHus2019gossip}, but with a variation: considering a one-to-many scheme. The randomness is established by using two parameters: $T$, the probability of sharing, and $s$ the number of peers that would potentially receive the information.

Second, each node analyzes if the model to share (set of local prototypes) significantly increases the knowledge of these peers (set of prototypes of the selected neighbors) (line $6$). To measure this information gain, we apply the Jensen-Shannon distance (JSD)~\cite{lin1991jsdistance}, derived from the Kullback-Leibler Divergence, that provides a symmetric and bounded measure of the difference between two distributions. The model would be shared only if this distance is higher than a threshold value $Th_{JSD}$. 

In order to calculate the JS distance among two set of prototypes (local models), we
first obtain the probability mass functions (pmf) that describe each set of prototypes, denoted as $P$ for the local one and $Q$ for the peer's one. After that, the JSD distance is calculated using~\eqref{eq:jensen-shannon-distance} and~\eqref{eq:divergence}. We propose to obtain the pmfs needed to perform the comparison ($P$ and $Q$) using the following procedure. First, apply the Kernel Density Estimation (KDE) to each prototype set to obtain a fitted continuous approximation to the true pmfs. Then, and with the goal of having two comparable pmfs, rescale both pmfs at identical points:

\begin{enumerate}
\item Apply KDE to obtain a first approximation to the pmfs $P$ and $Q$. With this purpose, we use the Epanechnikov kernel~\cite{epanechnikov1969non, moraes2021epanechnikov}, chosen because of its efficiency and the smoothness of the resultant pmf, as follows:
\begin{equation}
  \hat{f}(x) = \frac{1}{m h} \sum_{i=1}^m K \left(\frac{x - X_i}{h}\right),
  \label{eq:pmf1}
\end{equation}
where $K(u) = \frac{3}{4}(1 - u^2)$, satisfying that $|u| \leq 1$; $h$ is the bandwidth determined by Scott's rule $h = n^{-1/(d+4)}$, $m$ is the number of data points used in the estimation, and $d$ is the dimensionality of the data (number of features of each distribution); note that both distributions have the same dimension.
         
\item Once a first approximation of $P$ and $Q$ have been obtained, we adjust both of them to be comparable by generating random uniform points across all dimensions to dynamically adapt the number of samples based on the dimensionality and value range of the data
\begin{equation}
    N_\text{points} = \left\lfloor B_\text{points} \cdot 2^{d/2} \cdot \left(\frac{T_\text{range}}{2 \cdot d}\right) \right\rfloor,
\label{eq:pmf2}
\end{equation}
where $B_\text{points}$ is the base number of points (the more number of points the more precise calculations, although it entails a higher computational cost), $d$ is the number of dimensions and     $T_\text{range}$ is the sum of the range (maximum $-$ minimum) across all dimensions. For the experimental analysis (Sect.~\ref{sec:results}), $B_\text{points} = 1.000$ to ensure the generated point density would properly cover all the range of data.

\item Finally, we evaluate and normalize the density of the generated points to obtain two probability mass functions that can be compared using the JSD distance, as indicated in~\eqref{eq:jensen-shannon-distance}. This step is perform according to:
\begin{equation}
    p_j = \frac{\hat{f}(u_j)}{\sum_{j=1}^J \hat{f}(u_j)}, \, \quad \sum_{j=1}^J p_j = 1.
\label{eq:pmf3}
\end{equation}
\end{enumerate}

To sum up, we first apply~\eqref{eq:pmf1}, \eqref{eq:pmf2} and~\eqref{eq:pmf3} to 
obtain two comparable probability mass functions $P$ and $Q$ that represent two sets 
of prototypes of two different nodes in the decentralized learning network. After that, 
it is possible to obtain the distance between the two of them, $JSD(P, Q)$ as indicated 
in~\eqref{eq:jensen-shannon-distance}. If this distance is large enough, higher than a
threshold $Th_{JSD}$, then it is considered the knowledge provided by one of the nodes 
is worthy to be shared with the other node. 

\subsection{Compressing the model size}
\label{sec:clustering}

\begin{algorithm}[tpb]
    \setstretch{1.0}
    \caption{DBSCAN-Based Prototype Clustering by Label}
    \label{alg:dbscan_prototype_clustering}
    \begin{algorithmic}[1]
        \Procedure{DBSCANLabelClustering}{$\text{prototypes}$, $\text{labels}$, $\text{limit\_size}$, $\text{target\_range}$, $\text{eps\_initial}$}
            \State Initialize variables and set initial epsilon value for each label
            \For{each label in labels}
                \While{not within target range for the label}
                    \State Perform DBSCAN clustering on prototypes of the current label
                    \State Calculate the number of clusters (prototypes)
                    \If{number of clusters within target range}
                        \State \textbf{break}
                    \EndIf
                    \If{number of clusters $>$ target max}
                        \State Increase epsilon
                    \ElsIf{number of clusters $<$ target min}
                        \State Decrease epsilon
                    \EndIf
                    \If{number of iterations exceeds limit}
                        \State \textbf{raise} error
                    \EndIf
                    \State Update iteration count
                \EndWhile
                \State Update prototypes with new centroids from clusters
                \State Adjust relevance and neighbors for new prototypes
            \EndFor
            \State \Return final epsilon values for each label
        \EndProcedure
    \end{algorithmic}
\end{algorithm}

The ILVQ algorithm tends to increase the number of prototypes of the learning model, which is an important issue when dealing with resource-limited devices. In order to control this growth, we propose to use a clustering algorithm ({\texttt{CompressModelSize}}: line 6 in Algorithm~\ref{alg:learn_from_queue}) that  groups prototypes to a smaller number of prototypes, i.e., to compress the local model before sharing this information within the learning network. Thus, we have included this mechanism in the {\texttt{LearnFromPeers}} algorithm (line $6$ in Algorithm~\ref{alg:learn_from_queue}), which is run if the prototype size exceeds a threshold $Th_{prot}$. This solution optimizes resource usage, maintains system responsiveness, prevents from overfitting and reduces the communication overheads.

After analyzing several options, we have decided to use the DBSCAN algorithm that would detect and group similar prototypes, without excluding minorities. Once the clusters are detected, we propose a mechanism to merge those prototypes in the same cluster to obtain only one prototype, which would properly represent the complete set. This is done by calculating the centroid of each cluster by averaging the attribute and label coordinates of all prototypes within the cluster. The relevance scores, which measure the number of successful predictions, are summed across all prototypes within the cluster. This cumulative relevance score is then assigned to the centroid, effectively capturing the predictive effectiveness of the entire cluster. Once clustering is complete, the neighborhood for each new prototype is reconstructed by identifying and assigning the two closest prototypes within the updated set. This procedure, detailed in Alg.~\ref{alg:dbscan_prototype_clustering}, ensures that the local topological relationships are maintained in the new, reduced prototype set.

\section{Analysis}
\label{sec:analysis}

This Section presents an analysis of the dynamics of the gossip algorithm used
in our system. Our goal is to study the Age of Information (AoI) for this
specific gossip protocol, and lower bound the freshness of the model computed
locally at the nodes.

We assume that each node receives from its environment a sequence of
real-valued samples following an exponential time distribution between
arrivals (a Poisson process) with rate $\lambda_i$
$X_i \sim \text{Exp} (\lambda_i)$ for all $i \in [N]$. After receiving a new
sample, node $i$ trains its local model and then shares the updated model with
a random subset of its neighbors. To this end, node $i$ decides first whether
to share or not, with probability $T_i$, and if the action is to share, then
it chooses randomly a subset of $s_i$ neighbors to send them an
update. Hence, $s_i / (N - 1)$ is the fraction of neighbors that obtain a copy
of the updated model, where $n$ denotes the total number of nodes in the
system. For simplicity, we assume in the rest of this Section that
$\lambda_i = \lambda$, $T_i = T$ and $s_i = s$ for all $i$.

Accordingly, the local model at node $i \in [N]$ is updated whenever a new
local sample is read or when a new set of prototypes is received from some of
its neighbors. These combined arrivals follow a Poisson process with rate
$\lambda + \sum_{i = 1}^{N - 1} \lambda T s \overline{L} / (N - 1) = \
\lambda (s T \overline{L} + 1)$, since the gossip messages are sent uniformly to 
$s$ neighbors at random, and $s T$ is just the average number of copies sent out 
by a given node upon finishing its model update; $\overline{L}$ is the mean number 
of prototypes shared in every message, this is important because the prototypes will 
be processed individually, not the whole set at once. If $\overline{L}$ cannot be 
precisely calculated or estimated, the maximum size of the prototype set just prior 
to clustering (which reduces its size) could be used as a conservative estimate. 
Thus, we can model each node as a simple M/M/1 with arrival rate 
$\lambda (s T \overline{L} + 1)$ and service rate $\mu$, and this queue is stable 
when $\lambda (s T \overline{L} + 1) / \mu < 1$, or $s T \overline{L} < \mu / 
\lambda - 1$, which sets the upper threshold for the product $s T$. Nevertheless, 
a more realistic model for the input queues in our system to a node is that of a 
collection of $N - 1$ logical queues ---one for each neighbor---served in round-robin
order. Hence, we can better model this policy, at a first approximation, as a
single-server queue with GPS (Generalized Processor Sharing) scheduler. Since
our gossip protocol is completely homogeneous, the global bound for stability
is the same, $\lambda (s T \overline{L} + 1) / \mu < 1$. We summarize the 
result in the following lemma.

\noindent\textbf{Lemma 1}. For the randomized $(s, T)$-group gossiping
protocol, with exchange rate $\lambda$ and service rate $\mu$, the input
queues at the nodes are stable if $\lambda (sT \overline{L} + 1) < \mu$.

In addition to finding the condition for stability in the gossiping
process, we are interested in calculating the freshness of the gossip
messages, or equivalently the AoI, the age of the latest message
received at node $i$ from node $j$. For this, suppose that at each
epoch when a node updates its local model, a local counter (a logical
clock) is increased by one unit. Each gossip message transmitted by
node $i$ is labeled with the value of $i$'s logical clock, and we
define $N_j^i(t)$ as the version of model $i$ at node $j$ at time
$t \in \mathbb{R}^+$. Clearly, the stochastic counting process
$N_j^i(t)$ is a Poisson process because the time between updates
follows an exponential distribution. Further, we define the effective
update rate $\lambda_u$ at an arbitrary node as follows
\begin{equation}
  \label{analysis:ur}
  \lambda_u = \min \{ \lambda (s T \overline{L} + 1), \mu \},
\end{equation}
depending on whether the system is congested or not.

Further, we define the staleness or age of the message/model of $j$ at $i$,
denoted as $S_j^i(t)$, as the difference in the version numbers at two
nodes, i.e., $S_j^i(t) = N_i^i(t) - N_j^i(t)$. This represents the
number of versions by which the model at node $i$ is outdated compared
to node $j$. In the following lemma, we derive an upper bound for the
staleness of the decentralized learning system.

\noindent\textbf{Lemma 2}. The expected staleness of a user is bounded
by
\begin{equation*}
  \mathbb{E}[ S_j^i(t)] \leq \frac{\mu}{\lambda s T} \sum_{k = 1}^{N - 1} \frac{1}{k}.
\end{equation*}
\textit{Proof}: This is a direct application of~\cite[Lemma
1]{mitra2024scale}, replacing $\mu_i$ with $\mu$ ---since we assume
that all the nodes are identical, and $\lambda_i$ with
$\lambda s T$, since the arrival rate between a specific pair
of nodes is just $\lambda sT / (N - 1) $. 

An immediate consequence of this Lemma is a lower bound to the
gossiping capacity $\lambda$ of an individual user so that the
expected staleness is bounded. The following Lemma states this result.

\noindent\textbf{Lemma 3}. If the gossiping rate of an individual user
in a complete graph scales as $\Omega(\log N)$, the expected staleness
is bounded.

\textit{Proof}: The harmonic series can be approximated as
\begin{equation*}
  \sum_{k = 1}^{N - 1} \frac{1}{k} = \log (N - 1) + \gamma + O\left(
    \frac{1}{N} \right)
\end{equation*}
where $\gamma$ is the Euler constant. Taking $\lambda \sim \Omega(\log
N)$ in the bound of Lemma 2, we get
\begin{equation*}
  \frac{\mu}{\Omega(\log N)} \left( \log (N - 1) + \gamma + O\left(
      \frac{1}{N} \right) \right) = O(1).
\end{equation*}
Therefore $\lim_{n \to \infty} \lim_{t \to \infty} \mathbb{E}[S_j^i(t)]$ is a 
constant, implying that the expected staleness is finite in the steady state.

We remark here that this lower bound is the same as for
the gossiping protocol used in~\cite{mitra2024scale}, and the scaling
law coincides. This can be explained by observing that updating in parallel 
multiple nodes' local models  reduces the AoI by a factor $s$, which is constant 
in our case, so the same scaling remains valid. The results presented in the above 
Lemmas can be easily adapted for a variable number of neighbors updated 
simultaneously, i.e., for $s = O(N^a)$ for some $a > 0$. An important difference, 
however, with~\cite{mitra2024scale} is that in our modified gossiping the
staleness at different nodes is \emph{correlated}, since the updates
are sent simultaneously to a subset of neighbors. Thus, it is an
open question to determine, for our system, how the
correlated mixing of the local models conducted at different nodes
affects the global convergence of the model, i.e., whether a speedup
or a slow down happens due to this. It is also left for study the
empirical validation of this scaling, to show that it is possible to
achieve a linear speedup of the convergence with $n$, as found in
other similar studies.

\section{Validation and discussion}
\label{sec:validation}

In order to validate our proposal, we have deployed an experimental setup with five
Raspberry Pi 4 Model B+ units~\cite{raspberrypi2023} with 8GB RAM and 64GB storage,
interconnected via Wi-Fi and running the Ubuntu Server 22.04. Each one runs the ILVQ
algorithm and represents one of the nodes in the learning network 
(Fig.~\ref{fig:architecture_and_notations}). We use a Poisson process with rate 
$\lambda_s = 10$ samples per second to emulate the sensor data samples arrival and we 
integrate the ZeroMQ communication framework, particularly suited for IoT 
applications~\cite{akgul2013zeromq}, to facilitate efficient data transfers among nodes
over the Wi-Fi network. ZeroMQ architecture supports a scalable, many-to-many 
communication framework where nodes can bind to multiple receivers, and receivers can 
accept connections from multiple emitters. Additionally, its control system over 
message  delivery minimizes latency and optimizes network usage. Finally, messages 
are  serialized using the Python \texttt{pickle} module.

\subsection{Experimental tests}
\label{sec:test-definition}

Under this setup, we have firstly performed a baseline test ({\bf BaseTest}) without 
any optimizations or improvements, just the decentralized ILVQ algorithm detailed
in~\cite{gonzalez2024decentralized}. To compare the impact of each one of the proposed
improvements, we conducted our tests in the following four scenarios (i) the efficient
sharing mechanism based on the JSD ({\bf JSDTest}); (ii) the queue size limit of 
$10,000$ prototypes ({\bf LimitQueueSizeTest}) and (iii) the managing model size mechanism, 
that includes the DBSCAN algorithm, the random sampling and the LIFO queues, where only 
one set of prototypes are kept per neighbor ({\bf ClusteringTest}). We repeated 
$50$ independent runs of each combination of parameters in each one of the four 
scenarios in order to obtain a representative set of results.

For the sensing data, we have used the following two datasets available at the River 
online learning library~\cite{river2024}, which were selected because of their 
continuous and dynamic data streams for simulating real-world IoT conditions. The 
first one, the \texttt{Phishing Dataset}, is suitable for binary classification with
$1,250$ data points to distinguish legitimate from phishing websites. We set $R = 0$
to maintain uniformity in testing under static conditions. The second one,
\texttt{Electricity Dataset}, is a time-series forecasting task related to electricity
demand and pricing. This second dataset is used to test the data handling capacity of 
the model in two ways: (i) $R = 0$ and $D = 5{,}000$, assessing performance over a 
large sequential dataset, referred as \texttt{Electricity\_fixed Dataset} in this 
document; and (ii) $D = 1{,}250$ and random $R$ in each iteration, testing adaptability 
to varying data segments, referred as \texttt{Electricity\_random Dataset} in this 
document.

In order to split the data among the five nodes and each one to have equitable and
representative distribution of data, we applied the following mechanism 
$\mathcal{I}_m = \{ R + i \cdot N + m \mid i = 0, 1, \ldots, S-1 \}$, where $i$ 
is the sample index, $S$ is the number of samples per model, $N$ is the total number 
of models, $D$ is the dataset size, and $R$ is the starting index. Finally, we 
thoroughly assess the efficacy and efficiency of our proposal according to the following 
performance metrics:
\begin{itemize} 
\item F1 Score, which balances precision and recall, providing a measure of accuracy. 
It is calculated as: $F_1 = 2 \cdot (\text{precision} \cdot \text{recall}) / 
({\text{precision} + \text{recall}})$.
       
\item Average Number of Trained Prototypes per Node, which tracks the average number 
of prototypes each node trains during the experiment, providing insight into the learning 
activity and efficiency within the network.
        
\item Average Bandwidth Used per Node, which measures the average data transmitted per 
node during the learning process, excluding protocol overhead. It enables a comparison 
of data transmission efficiency across different scenarios, helping optimize 
communication costs and scalability in resource-constrained environments.
\end{itemize}

\subsection{Results}
\label{sec:results}

\begin{figure}[tpb]
    \centering
    \includegraphics[width=\textwidth, height=1.5\textwidth, keepaspectratio]{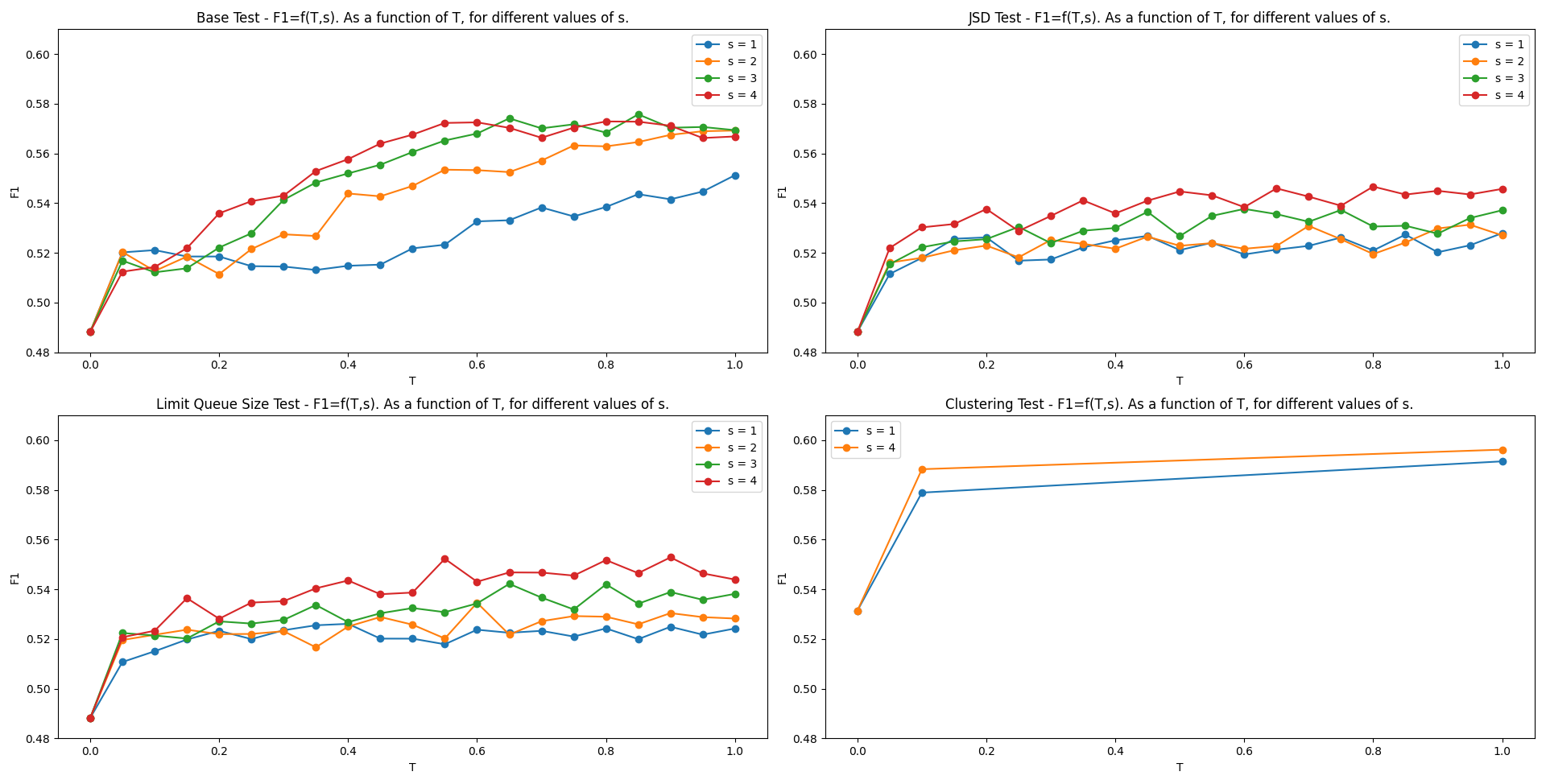}
    \caption{Performance comparison according to F1 Score (\texttt{Electricity\_fixed Dataset}).}
    \label{fig:F1score}
\end{figure}

In this Section, we summarize the obtained results for the three parameters 
previously enumerated: F1 Score, Average number of trained prototypes per node and 
Average bandwidth used per node. We have performed the tests using the three datasets 
(\texttt{Electricity\_fixed Dataset}, \texttt{Electricity\_random Dataset} and 
\texttt{Phishing Dataset}) and the four scenarios: {\bf BaseTest}, {\bf JSDTest}, {\bf 
LimitQueueSizeTest} and {\bf ClusteringTest}. However, it should be remarked that 
the results for the latter ({\bf ClusteringTest}) are limited to one dataset 
(\texttt{Electricity\_fixed Dataset}) and two sharing parameters ($s=1$ and $s=4$). 
These results are significant to see the improvement. 

\begin{figure}[tpb]
\centering
\includegraphics[width=\textwidth, height=1.5\textwidth, keepaspectratio]{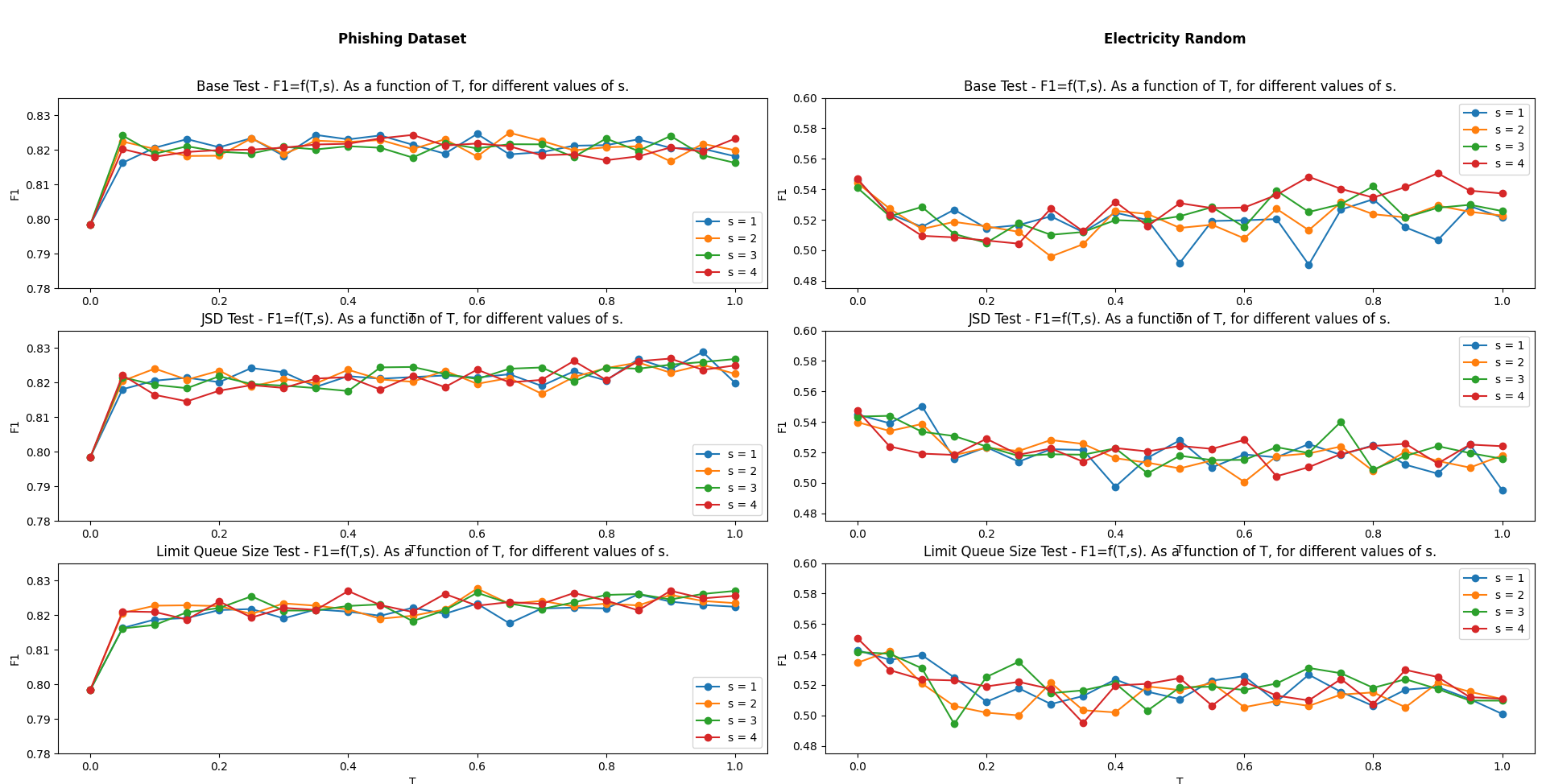}
\caption{Performance comparison according to F1 Score (\texttt{Electricity\_random Dataset} and \texttt{Phishing Dataset}).}
\label{fig:F1score_two_data}
\end{figure}

Fig.~\ref{fig:F1score} summarizes the obtained results for F1 Score using the 
\texttt{Electricity\_fixed Dataset} and Fig.~\ref{fig:F1score_two_data} shows the 
results for F1 Score using the other two datasets (\texttt{Electricity\_random Dataset} 
and \texttt{Phishing Dataset}). According to this information, we may conclude the F1 
Score increases when the enhancement mechanisms are tested ({\bf JSDTest}, 
{\bf LimitQueueSizeTest} and {\bf ClusteringTest}) compared to the {\bf BaseTest}. 
It is substantial the improvement in the {\bf ClusteringTest}: for instance, with
parameters  $T=1.0$ and $s=4$ and a limit of $500$ prototypes, the relative 
improvement of F1 Score when compared to the {\bf BaseTest} is almost a $20\%$, 
showing notable gains in model efficiency and accuracy. When reducing the number 
of prototypes to $50$, the gain is also reduces, but, again, F1 Score shows slightly 
better results. It is worth observing  that the general low performance results 
for F1 Score are consequence of using datasets with concept drift and the sampling 
method to divide them among the five computing nodes, which is a challenge for each 
node to accurately track the concept drift. However, and since the objective is 
assessing the improvement obtained with the proposed mechanisms and not he 
standalone performance of the model, we consider the assessment is still relevant. 
Anyway, when samples are taken consecutively, the model reaches a $75\%$ F1 Score.

\begin{figure}[tpb]
\centering
\includegraphics[width=\textwidth, height=1.5\textwidth, keepaspectratio]{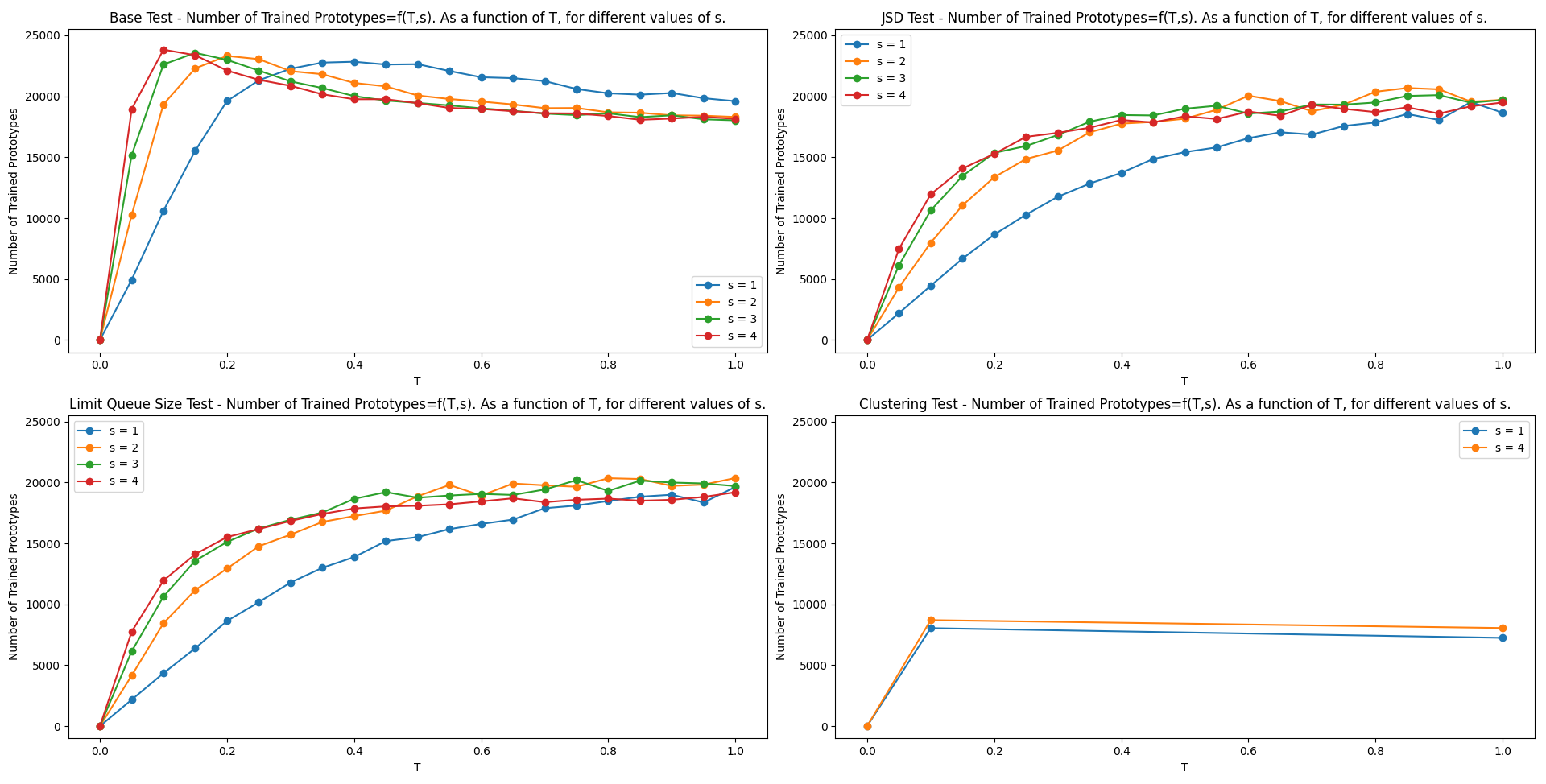}
\caption{Performance comparison according to the number of prototypes trained  (\texttt{Electricity\_fixed Dataset}).}
\label{fig:prototypes}
\end{figure}

\begin{figure}[tpb]
\centering
\includegraphics[width=\textwidth, height=1.5\textwidth, keepaspectratio]{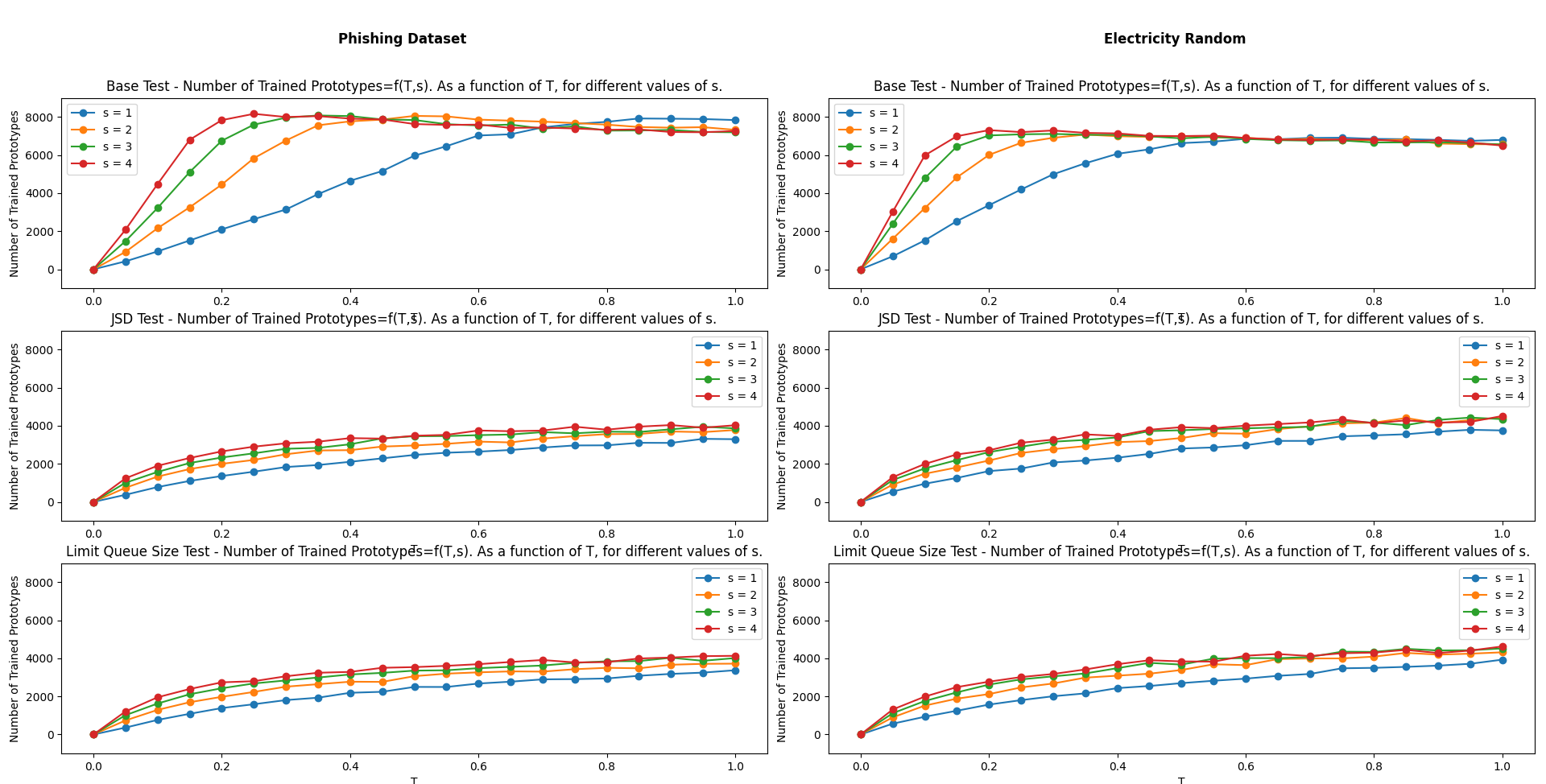}
\caption{Performance comparison according to the number of prototypes trained (\texttt{Electricity\_random Dataset} and \texttt{Phishing Dataset}).}
\label{fig:performance_two_data}
\end{figure}

Regarding the number of prototypes trained, Fig.~\ref{fig:prototypes} and 
Fig.~\ref{fig:performance_two_data} depict the obtained results in the same scenarios. 
When applying the limit to the queue size, {\bf JSDTest}, the saturation in the 
{\bf BaseTest} is controlled, showing a good behaviour of the selective sharing 
mechanism. Setting a limit on the queue size, {\bf LimitQueueSizeTest}, does not show 
a significant improvement. Finally, the {\bf ClusteringTest} shows a relevant inverse
correlation between the clustering limit size and the number of prototypes trained. 
This analysis reveals that higher limit sizes lead to less prototypes trained due to 
the higher computational overhead inherent to clustering. Thus, more resources assigned 
to clustering entails less resources available to train. It is particularly noteworthy 
that even with a limit of $50$ prototypes, the model trains more prototypes than the 
{\bf BaseTest}. This may be explained by the reduced computational demand of each 
training and prediction cycle in a distance-based algorithm, where fewer distance
calculations are required, making each operation cheaper and faster.

\begin{figure}[tpb]
\centering
\includegraphics[width=\textwidth, height=1.5\textwidth, keepaspectratio]{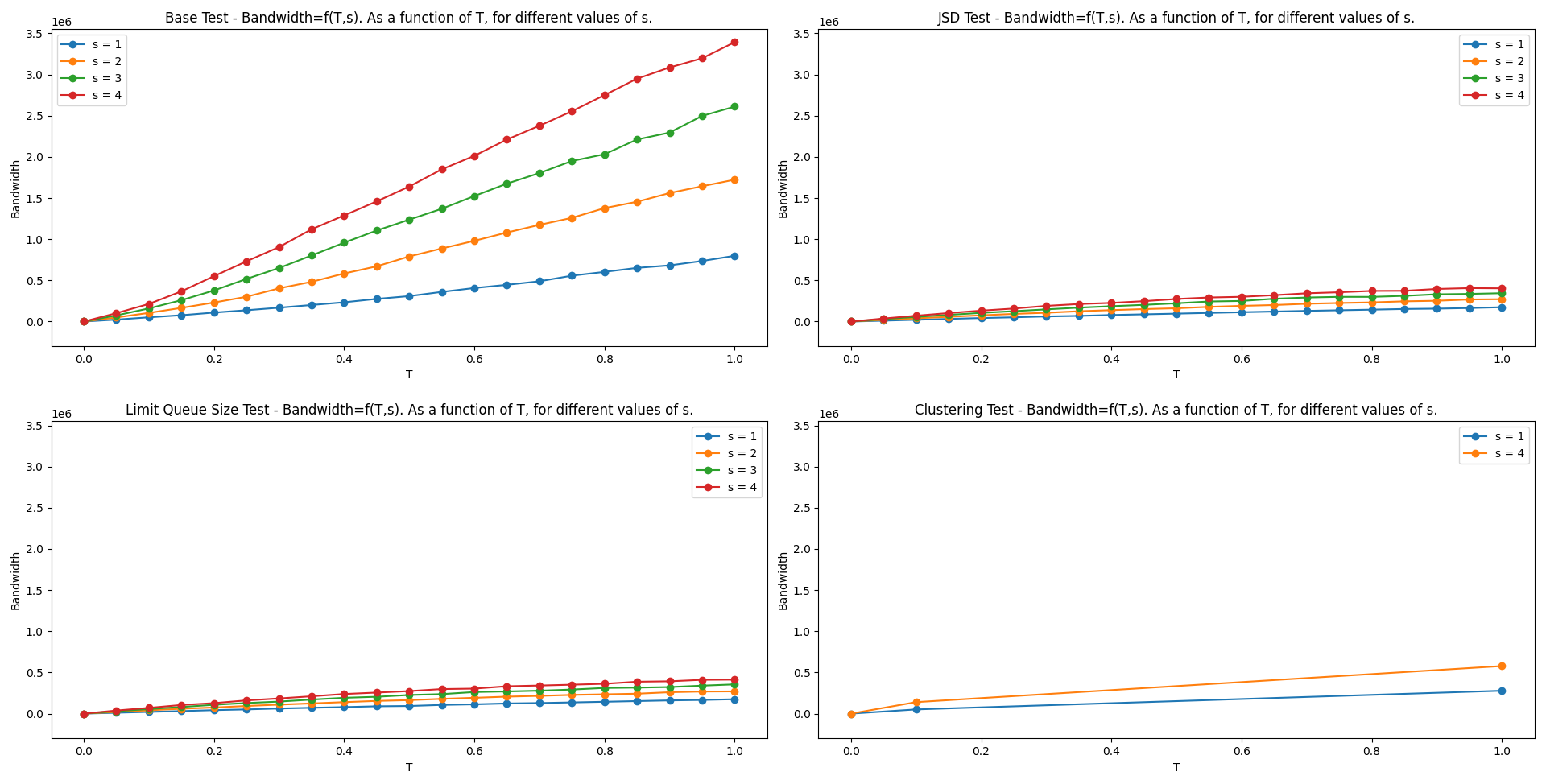}
\caption{Performance comparison according to the bandwidth usage  (\texttt{Electricity\_fixed Dataset}).}
\label{fig:bandwith}
\end{figure}

\begin{figure}[tpb]
\centering
\includegraphics[width=\textwidth, height=1.5\textwidth, keepaspectratio]{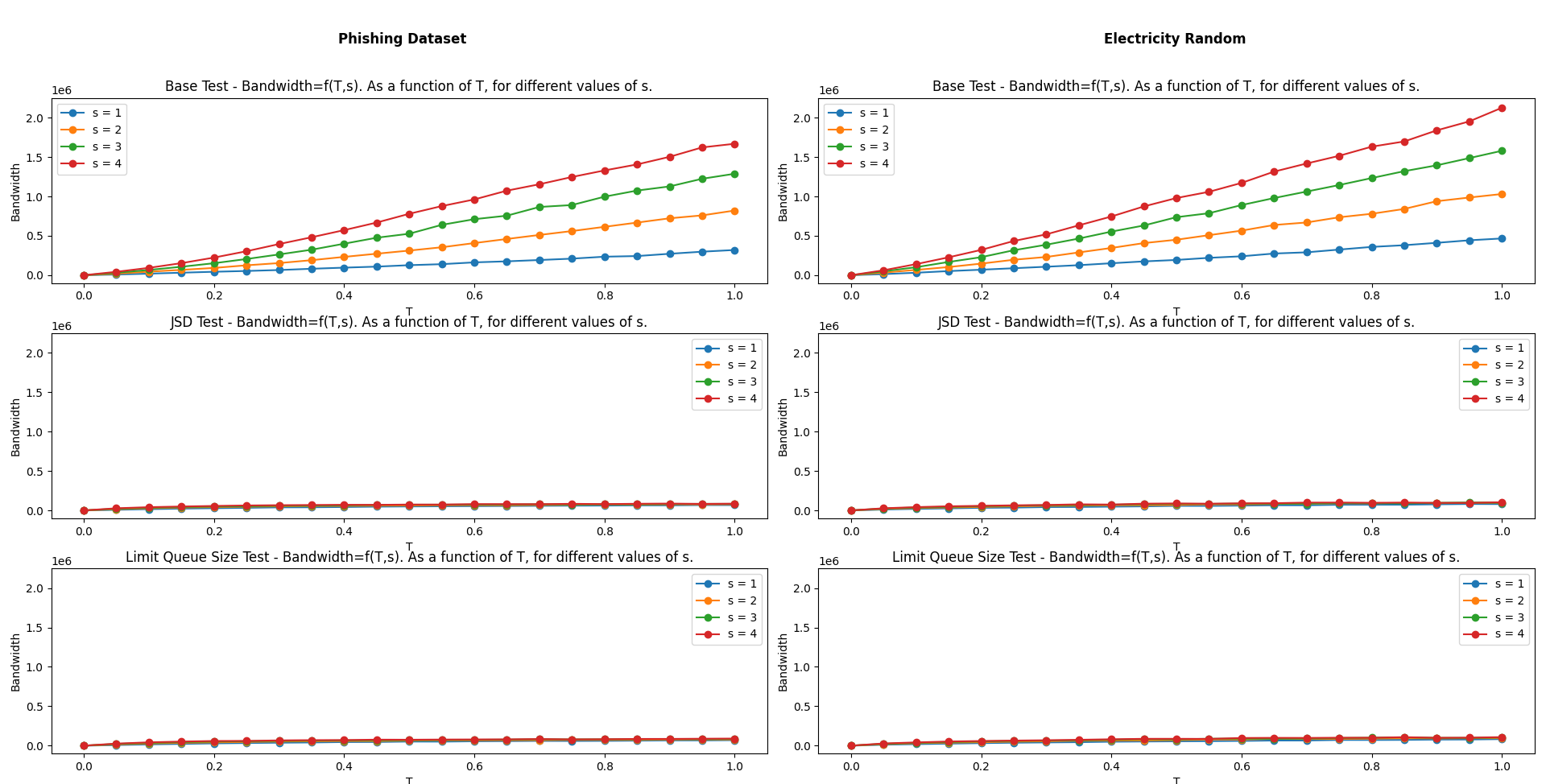}
\caption{Performance comparison according to the bandwidth usage (\texttt{Electricity\_random Dataset} and \texttt{Phishing Dataset})}
\label{fig:bandwidth_two_data}
\end{figure}

Finally, Fig.~\ref{fig:bandwith} and Fig.~\ref{fig:bandwidth_two_data} show the 
behavior of the proposal regarding the average bandwidth usage. Whereas it seems 
to see a linear pattern in the {\bf BaseTest}, the impact of the selective sharing
mechanism is clear in the {\bf JSDTest}. First, this usage is nearly constant for 
the different values of $s$. Second, the the data shows a dramatic reduction in 
bandwidth consumption from MB/s to KB/s. This would be understood as a good 
mechanism to minimize the sharing redundant information, precisely what we pursue in 
the first place. This behavior is kept in the {\bf LimitQueueSizeTest}, which is 
coherent with this improvement: the maximum queue size does not affect the exchange 
of models behavior, only the internal processing of information. The 
{\bf ClusteringTest} shows interesting behaviors. For lower values, such as $T = 0.1$ 
and $s=1$, most configurations use more bandwidth except for the $50$ prototype limit,
which uses less than half compared to the previous test. The $150$ prototype limit uses
slightly more, the $250$ limit almost doubles, and the $500$ limit more than doubles 
the bandwidth usage. However, this effect diminishes when higher values of sharing
parameters are used. Particularly, for $T=1$ and $s=4$ and $250$ and $500$ limits, 
the bandwidth usages increases about $30\%$ compared to the {\bf LimitQueueSizeTest}.
However, this increase is balanced with a notable improvement in performance, showing 
the efficiency trade-offs involved.

\subsection{Discussion}
\label{sec:discussion}

The previous results show, as expected, a clear dependency between the bandwidth usage, 
the number of prototypes processed, and the F1 score: (i) higher bandwidth usage 
generally led to better model performance and (ii) using a high prototype limit (such 
as $500$) entails more bandwidth needs, but achieves higher F1 scores. This highlights 
the relevance of an appropriate trade-off in these kind of decentralized learning 
settings between bandwidth usage (communications overhead) and model effectiveness.
Based on this observation, we can measure the network efficiency as a new parameter 
that relates the obtained performance (F1 Score) and the bandwidth usage. The analysis 
of this ratio offers interesting conclusions when we compare the {\bf BaseTest} and 
the {\bf ClusteringTest}.

The best results in the {\bf BaseTest} ($57\%$) is achieved with a bandwidth usage 
of $3.5$ MB/s, which entails an efficiency ratio of $16.14\%$ F1 Score per MB/s. 
The best results in the {\bf ClusteringTest} ($59.5\%$) is obtained with an
accumulated bandwidth usage of $305$ MBps and a prototype set limitation of $500$ 
and a parameter range between $72.5\%$ and $77.5\%$, which entails an efficiency ratio 
of $119\%$ per MB/s. This represents a \textbf{$737\%$ increase} in efficiency from the 
baseline, or \textbf{$7.37$ times higher} in efficiency, which shows a significant
optimization in resource usage while improving model performance, i.e., a good balance 
between resource usage and performance. 

These conclusions are relevant to deploy decentralized machine learning models 
within resource-constrained environments. They emphasize the need to carefully 
adjust prototype limits and sharing parameters to maximize the performance of 
decentralized learning systems while managing network resources efficiently.

\section{Conclusions and future work}
\label{sec:conclusions}

Totally decentralized learning settings are getting more attention, specially for 
resource-limited devices, such as in IoT solutions. In scenarios where the dynamical
adaptation to concept drift is important, incremental prototype-based learning 
algorithms offer an appropriate solution. Therefore, and taking into account our 
previous research works, we focused our attention on this kind of scenarios where 
the exchange of intermediate models among the computing nodes in the learning network 
is based on a gossip strategy.

Our proposal improves the efficiency of these collaborative schemes by
simultaneously tackling the following aspects. First, reducing the frequency of 
the exchange of local models within the learning network. With this aim we propose 
to share the intermediate model with a peer node only if the distance between both
intermediate models is significant enough to justify the exchange of information. 
We have opted to defined a solution based on the Jensen-Shannon Distance to assess 
this difference. Second, reducing the size of the intermediate model by defining a
clustering mechanism that compress the number of significant prototypes that 
characterize each local model. Finally, we have defined a scheduler (run at each node) 
that manages the incoming data to feed the prototype-based algorithm. This procedure 
tries to optimize the local learning process by correctly selecting those prototypes 
with the most recent and relevant information.

In order to assess these contributions we have applied two complementary procedures. 
On the one hand, we mathematically analyzed the dynamics of our adaptation of the 
gossip algorithm to obtain a lower bound for the freshness of the models computed 
locally at the nodes. On another hand, we have performed different experimental tests 
using resource-limited devices to measure the impact of our proposal. The latter has
demonstrated that our approach reduces the traffic in the learning network and the 
amount of information kept in each computing node without reducing the model performance. 

We are currently exploring three lines of work. One possible extension is analyzing
the use of adaptive model compression strategies that dynamically adjust the number 
of prototypes shared with peers and/or kept to be analyzed based on network conditions 
or convergence metrics. Another enhancement could be incorporating fault-tolerant 
mechanisms, such as self-healing protocols or distributed checkpointing systems, 
that would also be crucial to handle node failures and ensure continuous learning 
in dynamic and resource-constrained environments. Finally, we are assessing the 
possibility of combining a distributed stochastic bandit algorithm to improve the 
gossip protocol and get a more efficient use of the network resources.


\section*{Acknowledgments}

This work was supported by the grant PID2020-113795RB-C33
funded by MICIU/AEI/10.13039/501100011033 (COMPROMISE project), the grant PID2023-148716OB-C31 funded by MCIU/AEI/10.13039/501100011033 (DISCOVERY project). Additionally, it also has been funded by the Galician Regional Government under project ED431B 2024/41 (GPC).

\bibliographystyle{ieeetr}


\end{document}